\documentclass[letterpaper]{article} 
\usepackage{aaai25}  
\usepackage{times}  
\usepackage{helvet}  
\usepackage{courier}  
\usepackage[hyphens]{url}  
\usepackage{graphicx} 
\usepackage{natbib}  
\usepackage{caption} 
\usepackage{algorithm}
\usepackage{algorithmic}

\usepackage{makecell}
\usepackage{amsmath}
\usepackage{subfig}
\usepackage{newfloat}
\usepackage{listings}
\usepackage{multirow}
\DeclareCaptionStyle{ruled}{labelfont=normalfont,labelsep=colon,strut=off} 
\floatstyle{ruled}
\newfloat{listing}{tb}{lst}{}
\floatname{listing}{Listing}
\title{Automatically Planning Optimal Parallel Strategy for Large Language Models}
\author{
    Zongbiao Li\textsuperscript{\rm 1}\equalcontrib,
    Xiezhao Li\textsuperscript{\rm 1}\equalcontrib,
    Yinghao Cui\textsuperscript{\rm 1}\equalcontrib,
    Yijun Chen\textsuperscript{\rm 1},
    Zhixuan Gu\textsuperscript{\rm 1},
    Yuxuan Liu\textsuperscript{\rm 1},
    Wenbo Zhu\textsuperscript{\rm 1},
    Fei Jia\textsuperscript{\rm 1},
    Ke Liu\textsuperscript{\rm 1},
    Qifeng Li\textsuperscript{\rm 1},
    Junyao Zhan\textsuperscript{\rm 1},
    Jiangtao Zhou\textsuperscript{\rm 1},
    Chenxi Zhang\textsuperscript{\rm 1},
    Qike Liu\textsuperscript{\rm 1}
}
\affiliations{
    \textsuperscript{\rm 1}HUAWEI\\
    \{lizongbiao,lixiezhao,cuiyinghao,chenyijun31,guzhixuan,liuyuxuan30,zhuwenbo15,jiafei8,keke.liuke,liqifeng8,zhanjunyao,
    zhoujiangtao203,zhangchenxi30,liuqike1\}@huawei.com
}

\usepackage{bibentry}

\begin{document}

\maketitle

\begin{abstract}
The number of parameters in large-scale language models based on transformers is gradually increasing, and the scale of computing clusters is also growing. The technology of quickly mobilizing large amounts of computing resources for parallel computing is becoming increasingly important.
In this paper, we propose an automatic parallel algorithm that automatically plans the parallel strategy with maximum throughput based on model and hardware information.
By decoupling the training time into computation, communication, and overlap, we established a training duration simulation model. 
Based on this simulation model, we prune the parallel solution space to shorten the search time required.
The multi-node experiment results show that the algorithm can estimate the parallel training duration in real time with an average accuracy of 96\%. 
In our test, the recommendation strategy provided by the algorithm is always globally optimal.
\end{abstract}

%

\section{Introduction}

Scaling laws are driving large language models (LLMs) to become larger and larger in recent years\cite{kaplan2020scaling, hoffmann2022training}. The larger training data volume and larger models impose higher requirements on training hardware. Previously, engineers could train models on a single  Neural network Processing Unit (NPU), but now training large models requires multiple servers or even a large training cluster.

Collaborating such a large cluster to train a large language model is very delicate and complex.
Designers need to carefully consider how to allocate models to different NPUs and use extra ones to process data in parallel to accelerate training.
Many advanced distributed training methods, such as tensor parallelism\cite{dean2012large, shoeybi2019megatron}, pipeline parallelism\cite{huang2019gpipe} and data parallelism\cite{li2014scaling, li2014communication}, are proposed. 
Meanwhile, aiming to develop more scalable and efficient training processes, distributed training framework such as Megatron\cite{megatron_lm}, DeepSpeed\cite{rasley2020deepspeed} and ModelLink\cite{modellink} are designed. 
Combining multiple parallel strategies, these systems can train large models with billions of parameters on large clusters. However, for a large number of training hyperparameters introduced by multiple parallel strategies, these systems do not provide the basis for hyperparameter selection, but merely provide recommended empirical values. This makes it difficult for users to select hyperparameters for their own models.

The suboptimal parallel strategy can lead to increased training time, which means additional costs and is expensive in the development of high cost large-scale language models.
In the absence of guidance, users often need to do a lot of pre-experiments to determine a set of hyperparameters, which also means wasted time and increased costs.

Based on this dilemma, some work provides hyperparameter search strategies for these systems\cite{chen2024internevo, isaev2023calculon, miao2022galvatron}. 
However, due to the complexity of parallel frameworks, finding the globally optimal parallel strategy quickly and accurately remains a challenge.
In this article, we propose an automatic planning algorithm for finding the optimal parallel strategy.
Our algorithm first simulates the training duration, and by analyzing and modeling at the operator level, we can achieve an average estimation accuracy of 96\% for the training time. Then, based on the simulation model, we establish a pruning strategy that can prune 99\% of the search space, making it easy for us to enumerate the most efficient parallel strategies.

Unlike previous work, our algorithm covers a more comprehensive range of parallel hyperparameters, including terms such as micro batch size and global batch size that are often overlooked. 
The main content of this paper will be divided into the following three parts. 
\begin{itemize}
    \item Training duration simulation: We divide the parallel training duration into several sub items: computation, communication, and overlap, and simulate each items. Based on partial estimation, we can estimate the total training time required for each parallel strategy.
    \item Pruning and Searching: We first propose a complete planning model with a very large feasible range. Based on the simulation model, we prune the search space, reduce the feasible range size by nearly 99\%. Finally, we search for the optimal parallel strategy within a very small domain.
    \item Experiment and verification: We experiment with various Large Language Models on a small-scale cluster to verify whether our simulation algorithm and search algorithm are accurate. 
\end{itemize}



\section{Related Work}

\subsection{Basic parallel methods}

Each parallel strategy has its own focus and limitations. Therefore, efficient training of LLMs on large scale usually requires a combination of multiple parallelization methods.

\textbf{Data parallelism (DP)} can divides a large batch size of data to multiple workers\cite{li2014communication}. It replicates the entire model on multiple workers and accelerates the learning efficiency of the model through parallel computing. The respective gradient will be accumulated periodically to ensure the consistency of the weights in different workers. 

\textbf{Tensor parallelism (TP)} partitions weights and activation tensors of LLMs over multiple devices\cite{dean2012large}, and communicates at specific locations at each layer to aggregate block tensors. The parameters can be split along its row or column dimension to reduce the number of communications. With the segmentation strategy provided by Megatron-LM\cite{shoeybi2019megatron}, each transformer layer only requires two communications in the forward/backward computation. 
However, because TP communication occurs at each layer, the communication frequency is relatively high. In the backward stage, a certain overlap algorithm\cite{rashidi2021enabling, wang2022overlap} can be used to perform communication and calculation at the same time, thus reducing communication time cost.

\textbf{Pipeline parallelism (PP)}\cite{jia2019beyond,yang2021pipemare} splits the layers of LLMs into parts and allocates them to multiple workers. Activetion are communicated point-to-point between workers and are calculated in sequence. Because the execution of different layers of LLMs must wait in line, a certain amount of time waste is inevitable. This extra time is called pipe bubbles. There have been a lot of methods trying to reduce the pipeline bubbles. GPipe \cite{huang2019gpipe} divided a training batch into multiple micro batches and queued for computation to reduce the waiting time of subsequent NPUs. PipeDream-Flush \cite{narayanan2021memory} is an improved version of GPipe. It chooses the strategy that one forward pass followed by one backward pass (i.e., 1F1B), which effectively reduces the memory usage compared to GPipe. By dividing the reverse calculation into two parts, the new work\cite{qi2023zero} achieves lower bubble rate and effectively improves the throughput of PP.

\textbf{Others parallelism.} The combination of DP TP PP is called 3D parallelism. In addition, there are many other parallel ways.
Some methods solve the memory overhead introduced by LLM long sequences through sequence parallelism\cite{li2021sequence, korthikanti2023reducing}. 
Optimizer parallelism \cite{zero-infinity}divides optimizer states, gradients, and parameters to reduce redundant memory usage. This makes it possible to train larger models with limited memory resources.
Parallel method for Mixture-of-Experts (MoE) model is also proposed\cite{kim2021scalable}, and the model parallelism is carried out by splitting experts.
Most of these parallel approaches are independent of 3D parallelism and can be optimized separately. In addition, most distributed frameworks are based on 3D parallelism. Therefore, in this paper, we only consider the optimal strategy search of 3D parallelism, and will continue to incorporate other parallel methods in the future.



\subsection{Prior strategy search methods}

Before our work, there was some other projects tried to solve the problem of optimal parallel strategy searching.  

Mindspeed\cite{mindspeed} provides a strategy search algorithm based on profile, which can perform probabilistic search in the pruned strategy space. The profiling-based algorithm provides precise order-preserving estimation, but it takes a long time to search.
Gavatron\cite{miao2022galvatron} uses decision tree and dynamic programming to search for optimal strategy, and supports asymmetric model parallelism across devices. It only uses profiling for computing power estimation, training time is obtained by simulation, which greatly reduces the time of searching algorithm. However, it ignores the optimization of micro-batch. We demonstrate that ignoring this parameter may result in missing the global optimum.

As an analytical performance model, Calculon\cite{isaev2023calculon} provides a training time estimation method, which can also be used to select the optimal parallel strategy. In contrast to Gavatron, Calculon sets the micro batch size to the maximum size that the memory can hold for the sake of computing density. However, our experiments indicate that there is an optimal value for micro batch size. When the optimal value is exceeded, additional increase will only lead to an increase in bubbles, which in turn reduces training efficiency.
InternEvo\cite{chen2024internevo} is a new parallel training framework with specific optimizations for long sequence transformers. The training time estimation and automatic parallelism are included in the framework, but part of the communication is ignored, which may affect the order preservation.



Unlike previous work, our autoparallel algorithm takes into account more comprehensive variables, such as global batch size and micro batch size, which are often assumed to be constants in other work. The newly introduced variables will cause the search space to become larger, especially global batch size will expand the search space to infinity. In order to reduce the complexity caused by the increase of variables, we set up a white-box simulation system. The mathematical proof based on the white-box system helps us to prun the search space a lot, even if more variables are introduced, the search strategy space becomes very limited.

\section{Auto Parallelism Process}

\begin{figure}[htb]
  \centering
  \includegraphics[scale=0.325]{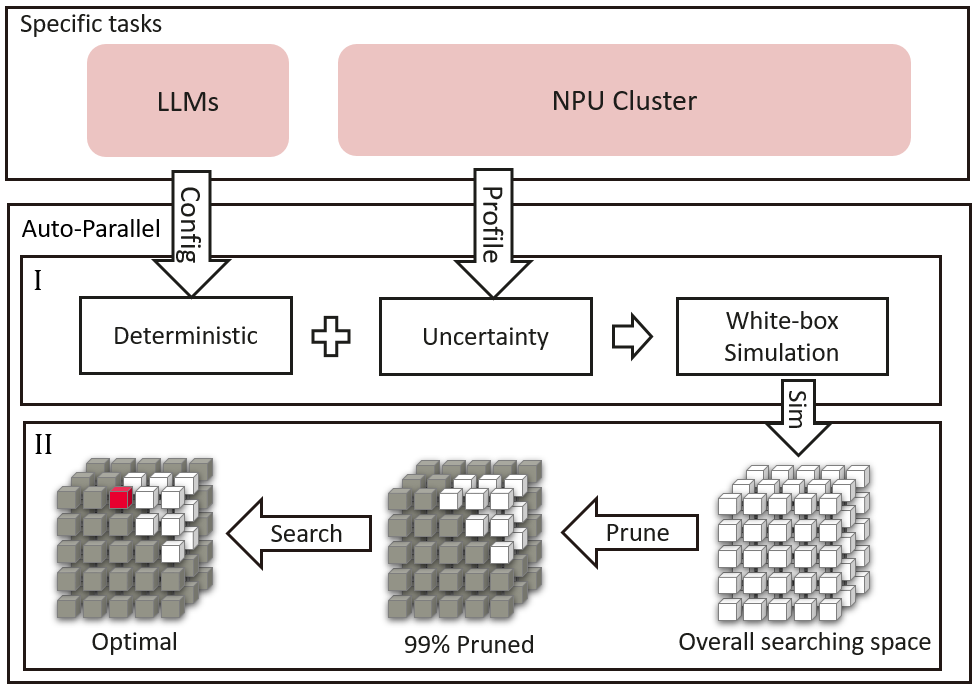}
  \caption{The overall workflow of our automatic parallel algorithm.}
  \label{pdf}
\end{figure}

We use Figure \ref{pdf} to summarize the overall process of our auto-parallel planing algorithm. The implementation of the algorithm can be divided into two steps: training duration simulation and pruned search of the optimal strategy.

The first step is to establish a simulation model based on the information of the model and training cluster, which can be used to estimate the required training duration for each strategy.
The simulation of training duration can be divided into deterministic part and uncertain part. The deterministic part includes the computation volume, communication volume of the transformer model, which is static with the training test, and can be estimated by direct calculation. The uncertain part includes the utilization rate of computing power and effective transmission bandwidth of the cluster, which are dynamic during the training process and obtained through profiling. This profile can be completed in advance to ensure a quick start for each training task. Alternatively, it can be performed before each training session to reduce interference.

In the strategy search algorithm, due to the consideration of multiple parallel variables, the search space is large and the complete search will be impossible. Based on this, we have established a search space pruning method based on the simulation model. It is possible to narrow down the search scope by more than 99\%. Finally, we find the optimal strategy through enumeration in a relatively small search space. Next, we'll discuss two parts of the algorithm in detail.


\section{Training Time Estimation}
\subsection{Notation and target}

To ensure the effectiveness of optimal strategy search, the simulation of training duration must be accurate and stable, and the order-preserving property must be provided. 
In this section, we introduce our training duration simulation model. Table \ref{notation1} lists symbols to be used in this paper.

\begin{table}[ht]
	\caption{Notation used in paper.}
        \begin{tabular}{ll} \label{notation1} \\
        \hline
        \underline{\textbf{\emph{Duration division}}} \\ 
        $T_{T}$   & Total time for training a epoch \\
		$T_F$     & Time for computing  \\
		$T_{CT}$  & Time for TP communication \\
        $T_{CP}$  & Time for PP communication   \\
        $T_{CD}$  & Time for DP communication \\
        $T_{B}$   & Bubble time   \\
        $T_O$     & Overlapped communication time \\
        $T_{AT}$  & Time for TP all-reduce computation \\
        $T_{AD}$  & Time for DP all-reduce computation \\
        
		\underline{\textbf{\emph{Hardware Config}}} \\
		$n$ & Number of NPU  \\ 
        $g$ & Network bandwidth between servers \\
   	$g_2$ & Network bandwidth within servers \\
        $q$ & Communicate slow down rate \\
        $\rho$ & Computing power utilization (Re) \\
        $U_{max}$ & Max computing power per NPU \\
  
		\underline{\textbf{\emph{Model Config}}} \\  
        $s$   & Sequence length \\
		$h$   & Hidden layers \\
        $a$   & Num of attention heads \\
		$H$   & Feed-forward hidden size  \\
		$L$   & Number of layer\\
		$V$   & Vocabulary size   \\
		$S_{t}$   & Number of total samples\\
        $u$   & Size of each model parameter \\
		
		\underline{\textbf{\emph{Parallel Setting}}} \\ 
        $t$   & Degree of tensor parallelism  \\
		$p$   & Degree of pipeline parallelism  \\
		$d$  & Degree of data parallelism \\
        $G$  & Global batch size \\
        $B$  & Mini batch size \\
		$b$ & Micro batch size   \\
        $m$ & Micro steps   \\
        \hline
    \end{tabular}
\end{table}

In order to make an accurate simulation, we split the training duration into several items and model them separately. Specifically, the total training duration is represented as equation \ref{TT_divi},

\begin{equation} \label{TT_divi}
\begin{aligned}
    T_{T}= (T_F+T_{CT}+T_{AT}-T_O+T_{CD}+\\
    T_{AD}+T_{B}+T_{CP})\frac{S_{t}}{G}
\end{aligned}.
\end{equation}

In addition to the main computation and communication, we also consider the computing duration of all-reduce, which is a proportion of the communication duration. This item takes a relatively small proportion of time. However, for the accuracy of subsequent analysis, this item is still included. 
This part of the calculation is not provided in the paper, but can be obtained by multiplying the corresponding communication time by a computing power/bandwidth coefficient.

The parameters required for modeling include hardware configuration, model configuration, and parallel settings.
The hardware parameters are obtained based on the information of the training cluster.
The communicate slow down rate $q$ refers to a bandwidth decrease caused by 
communication jamming when the communication members increases. As analyzed by InternEvo \cite{chen2024internevo}, the increase of intra-node and inter-node communication members reduces the efficiency of communication operators such as all-reduce, reduce-scatter, and all-gather. Computing power utilization refers the actual computing power compare to its maximum, which change with the computing intensity of model. We use $\rho$ to represent its reciprocal for easy publicity.Parameters that affect the tensor shape, such as $t$ and $b$, affect the computing power utilization.

The model parameters are obtained based on the large language model that the user needs to train. The parallel parameters are objects that need to be optimized in this work and are regarded as the unknown.
Compared with other work focusing on automatic parallelism, we do not assume that data parameters such as global batch size and micro batch size are constant. In the experimental section, we will prove that these parts affect the solution of the optimal.


\subsection{Computation duration}

The computation duration is usually the largest part of the total parallel training, so the accuracy of the estimation is important. 
The process we do the simulation can be summarized as follows: 
\begin{itemize}
\item Computation duration = Number of floating point operations per NPU $\times$ Maximum computing power per NPU $\times$ Computing power utilization.
\end{itemize}

For the first part of the above equation, given that the dense large language model is a variant of the Transformer model, we estimate the number of floating point per NPU by analyzing the Transformer architecture. 
The compute of the transformer-based large language model is mainly from tensor multiplication, which happen mainly at the attention, feed-forward, and vocabulary output layer. The computation of the vocabulary output layer is proved to be unnegligible, because other devices have to wait for this part in pipeline parallel, which introduces a long extra bubble time.


The computing power utilization of the NPU is related to the operator implementation and hardware. Here, we treat it as a black box and modelling it by small-scale profiling. Specifically, computing power utilization $\rho$ is considered as a function of $b, s, h, t$. The simulation function of $T_F$ is shown in equation \ref{TF},

\begin{equation} \label{TF}
T_F = (\eta_1+\eta_2 p) G \rho,\\
\end{equation}

\begin{equation} \label{eta}
\left\{
\begin{aligned}
    &\ \eta_1 = \frac{2(1+k)L(4sh^2+2s^2h+2shH)}{nU_{max}} \\
    &\ \eta_2 = \frac{2(1+k)shV}{nU_{max}} \\
\end{aligned}
\right..
\end{equation}

The specific modeling process is not given in the paper, but it is worth noting that we are only simulating the last NPU of the PP stage, which contains the vocabulary output layer. 
If we do not simulate the last stage of the PP, the previous NPU must wait for the last stage to complete the computation of the output layer, which makes the bubble time more complex.

\subsection{Communication duration}

Similar to the simulation of the computation duration, the simulation of the communication duration is carried out by the following basic ideas:

\begin{itemize}
\item Communication duration = Total communication volume / (Full speed bandwidth $\times$ Slow down rate).
\end{itemize}

The total communication volume depends on the model and the parallel approach, and this part of the simulation is static and can be obtained through direct analysis of the parallel training framework.
On the other hand, the full speed bandwidth and bandwidth slow down rate are dynamic during the training process, depending on the physical connection form and communication algorithm.
For example, if a meshed connection and communication algorithm are used within the server, the bandwidth of communication will not decrease with the increase of members. On the other hand, in a ring communication method, when there are fewer communication members, the bandwidth will be high, but as the number of members increases, the bandwidth will linearly decrease.
Same as the estimation of computation duration, the dynamic part of communication is obtained through profiling.

\subsubsection{Data parallelism.}
DP communication occurs at the end of each step and is used to synchronize model weights across different devices. The communication methods employed here is ring all-reduce, the duration estimation is shown in equation \ref{Tcd},

\begin{equation} \label{Tcd}
\begin{aligned}
    T_{CD} =  \lambda_1 (\frac{1}{pt} - \frac{1}{n}) + \lambda_2 (\frac{1}{t} - \frac{p}{n}) ,\\
\end{aligned}
\end{equation}

\begin{equation} \label{ambda_}
\left\{
\begin{aligned}
    &\ \lambda_1 =  \frac{2L(4h^2+2hH+9h+H)u}{g} \\
    &\ \lambda_2 =  \frac{2Vhu}{g} \\
\end{aligned}
\right..
\end{equation}

\subsubsection{Tensor parallelism.}

The intra-server communication adopts in our work is the mesh grid method, which means that the all-reduce of TP can be one-to-all. Therefore, the TP communication duration based on this architecture merely increases with the TP degree. However, as the number of communication members increases, bandwidth congestion may occur, resulting in a slight increase in communication duration. We evaluate this effect using a slow rate $q$, $q$ obtained by profiling. 
The simulation for TP communication is presented in equation \ref{TCT},

\begin{equation} \label{TCT}
T_{CT} = 
\left\{
\begin{aligned}
& 0, &t=1\\
& \gamma_1 + \gamma_2 p, &t > 1 \\
\end{aligned}
\right.,
\end{equation}

\begin{equation} \label{gama}
\left\{
\begin{aligned}
&\ \gamma_1 =  \frac{8LshuG}{qg_2n} \\
&\ \gamma_2 =  \frac{shuG}{qg_2n} \\
\end{aligned}
\right..
\end{equation}

For accuracy, we model the last worker of the PP queue, which introduces additional vocabulary communication.

\subsubsection{Pipeline parallelism.}

The time consumption caused by PP consists of two parts: bubble and point-to-point(P2P) transmission.

Bubble is the main time-consuming part, since NPU at the end of the queue must wait for the preceding members to complete their computation and communication. Some interleaved pipeline scheduling can reduce this wait time and keep the NPU in compute\cite{huang2019gpipe, yang2021pipemare}. Here, we use the PipeDream-Flush\cite{narayanan2021memory} for modeling, also known as 1-forward 1-back (1F1B) scheduling. After an initial startup period, all NPUs enter the computing state. In addition, the activation cache does not increase infinitely with micro-steps. The simulation of bubble time is shown in equation \ref{TBP},


\begin{equation} \label{TBP}
\begin{aligned}
    T_{BP} =  \left ( T_{T} + T_{CT} + T_{ART} - T_{O} \right ) \times \frac{p-1}{m}
\end{aligned}
\end{equation}

It is important to note that bubble time refers to the waiting period for other NPUs to complete their computations and communications. Thus, both computation and TP communication contribute to bubbles.

Another time-consuming item is P2P communication. PP communication transfers an activation to the next NPU when its computation ends, which takes a relatively short time. The estimation of PP communication is presented in equation \ref{TCP},

\begin{equation} \label{TCP}
T_{CP} = 
\left\{
\begin{aligned}
& 0, &p=1\\
& \alpha p t + 2 \beta p b - 3 \beta b, &p > 1 \\
\end{aligned}
\right.,
\end{equation}

\begin{equation} \label{alpha_beta}
\left\{
\begin{aligned}
&\ \alpha =  \frac{shuG}{gn} \\
&\ \beta =  \frac{shu}{g} \\
\end{aligned}
\right..
\end{equation}


\subsection{Overlap}

In our training framework, all-reduce occurs in both the multi head attention layer and feed-forward layer. This communication can overlap with computation during back propagation.
So the basic idea of communication overlap is that, the communication of the activation and the computation of weight can be synchronized during the backward stage.

\begin{figure}[ht]
\centering
\subfloat[All-reduce of multi head attention]{\includegraphics[scale=0.5]{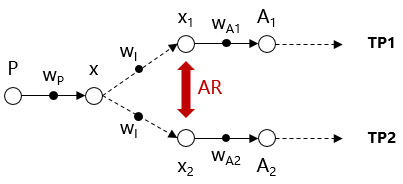}}\\
\subfloat[All-reduce of feed-forward]{\includegraphics[scale=0.5]{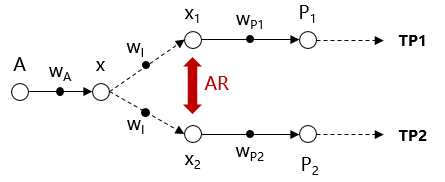}}
\caption{Position of communication overlap.}
\label{ar}
\end{figure}

Figure \ref{ar} shows the stages of two all-reduce operations, with the direction of the arrows indicating forward propagation. During backward propagation, the communication between $X_1$ and $X_2$ can be synchronized with the computations and updates of $W_{A1}$, $W_{A2}$, $W_{P1}$, and $W_{P2}$.

The time estimation of overlap can be summarize as equation \ref{overlap},

\begin{equation} \label{overlap}
\begin{aligned}
    &\ T_{O} = \min \Big ( T\big(AR(\frac{\partial L}{\partial x_1}, \frac{\partial L}{\partial x_2}), T(\frac{\partial L}{\partial w_L})\big) \Big )\\
\end{aligned}.
\end{equation}

The all-reduce communication time of activation has been described in the TP communication duration, and the computation time of the weight can be expressed as:

\begin{equation} \label{overlap_w}
\frac{\partial L}{\partial w_L}=
\left\{
\begin{aligned}
    &\ \frac{3ush^2L}{nU_{max}} G \rho , MAH\\
    &\ \frac{ushHL}{nU_{max}} G \rho , MLP\\
\end{aligned}
\right..
\end{equation}

There are two formulas, mainly due to the different weights that need to be updated in the multi-headed attention (MHA) layer and the multi-layer perceptron (MLP) layer.

\section{Strategy Searching}

\subsection{Overall searching space}

By simulating the training duration items, we be able to obtain the specific expression of Formula \ref{TT_divi}, thereby whitening the original black box training duration model. Next, we analyze the constituents of our simulation function and establish an integer programming problem as target:

\begin{align} \label{obj}
&\mathop{\arg \min}_{p,t,d,G,b,m} \quad T_{T}\\
& \begin{array}{r@{\quad}l@{}l@{\quad}l} \label{obj_st}
    s.t. \quad &p,t,d,b,m,G\in N^+\\
    &n=ptd , \quad 1\leq p \leq L, 1\leq t \leq 8\\
    &G=bmd\\
    &pt\times M_{NPU}\geq M_m \\
\end{array} .
\end{align}

$M_{NPU}, M_{m}$ refers to the memory capacity of each NPU and the memory required by the model. This constraint is used to indicate that the NPU memory meets the requirements of the model. This is a very naive representation, and we'll discuss this part later.

The complete objective function consists of the items in the simulation model. We will not expand this formula in the text, since it will take up too much space.
This programming function cannot be solved by brute-force, as it contain global batch size $G$ and micro batch size $b$, which is infinite in the search space. So in this chapter we prun the parameter space by memory restriction and mathematical analysis. With appropriate constraints, most of the search space, include $G$ and $m$ can be pruned by analysis.




\subsection{Search space pruning}

Unlike the degree of $p$ and $t$, which are limited by the number of NPUs, global batch size $G$ and micro batch size $b$ have unlimited values, which is the primary difficulty of the programming  problem.
We first prove that the total time is monotonic with respect to $G$, by rearrange function \ref{obj} as:

\begin{equation} \label{prof_gbs}
\begin{aligned}
    & T_T= (\phi_1+\phi_2 \frac{1}{G})S_{t} \\
\end{aligned}.
\end{equation}

Item $\phi_1, \phi_2$ is the rearrangement of the simulation function and is strictly positive. It is easy to see that $G$ has a multiplier effect on training time. When the $G$ is large, the time consumption items of $\phi_2$ is effectively reduced, and the $\phi_1$ part remains static. 
Therefore, from a perspective of throughput and training efficiency, a large $G$ is preferred. 
However, large $G$ also means less randomness for gradient, which can lead to a loss in model performance and impair the overall training efficiency. Based on this trade-off, the training of large language models should determine a global batch size according to the requirement of gradient randomness, but a larger global batch size can accelerate the training, by reducing the computation time, bubble time, and DP communication time in $\phi_2$


The pruning of micro batch size $b$ is generally based on two considerations: computing efficiency and memory limitation. In terms of computing efficiency, the increasing of $b$ expands the input tensor shape of each operator, which boost computing power utilization, but the increase in computing power is marginally decreasing. On the other hand, an increase in $b$ leads to an increase in bubbles. 
Based on this trade-off, we think that there is an optimal value for $b$, which is easy to verify by the partial derivative of $T_T$, as Formula \ref{prof_b}.

\begin{equation} \label{prof_b}
\begin{aligned}
    & \frac{\partial T_T}{\partial b} = \omega_1 \frac{\partial \rho(b)}{\partial b} + \omega_2 b\frac{\partial \rho(b)}{\partial b}+\omega_3 \rho + \omega_4 ,\\
\end{aligned}.
\end{equation}

Where $\omega_1,\omega_2,\omega_3,\omega_4$ are strictly positive. 
Also, since $\rho$ is inversely proportional to $b$, the sign of the partial derivative in the right expression of Formula 16 is negative. 
Therefore, there exists an optimal value for $b$. By addressing the relaxation problem of formula 16, we can determine a possible range of $b$.
This inference greatly pruns the search space and makes enumeration possible.

As we increasing $b$ for better computing power utilization, the size of the activation also increases, which greatly increases the memory pressure of parallel training. Moreover, a larger activation memory footprint means that we need to use more NPUs for model parallelism, reducing the resources available for data parallelism, which increases training time. 
Next, let's limit the size of b by analyzing the memory.



\subsection{Memory boundary}

One approach to pruning the search space is to prun the strategy of significantly exceeding the limit of the memory capacity.
In part of the work, memory capacity is considered a gray-box model. In actual scenarios, memory fragmentation and memory reclamation time points are unknown. As a result, the actual memory utilization may not be as expected.
However, a basic memory simulation model can still guide our work to some extent, pruning strategies that exceed the memory limit.

In object function \ref{obj_st}, we use naive formulas to describe memory limits.
Based on the naive idea, that the pipeline parallel and tensor parallel should meet the total memory required for model. Here we provide a more accurate description.
During the training duration, the NPU memory usage includes the model weight, optimizer status, and activation. 
The total memory usage during the training process is as follows:

\begin{equation}
\begin{aligned}
    M_{NPU} > \frac{M_m}{pt}\\
\end{aligned},
\end{equation}

\begin{equation}
\left\{
\begin{aligned}
    M_m &=M_w + M_{o} + M_a \\
    M_w &=L(4h^2+2hH+9h+H)u+Vhu \\
    M_{o} &=9M_w\\
    M_a &=Lsbp(18h+4H+5sa)
\end{aligned}
\right..
\end{equation}

$M_w, M_o, M_a$ indicate the memory usage of weight, optimizer and activation respectively.
Here, the 1F1B scheduling strategy is used to reduce the activation memory usage. The peak activation size that needs to be stored in the pipeline queue is reduced from $bm$ to $bp$.
Besides, note that the input activation are not tensor parallelized\cite{korthikanti2023reducing}, which further increases memory overhead.

Based on this, we can update the boundary of $p$ and $t$ to be:

\begin{equation}
\left\{
\begin{aligned}
    &\frac{m_2}{m_1} \leq t \leq 8 \\
    &\frac{m_3}{m_1 t-m_2} \leq p
\end{aligned}
\right.,
\end{equation}

where

\begin{equation}
\left\{
\begin{aligned}
    &m_1 = M_{\textit{NPU}}-2Lsb(h+H)\\
    &m_2 = Lsb(16h+2H+5sa)+Vhu\\
    &m_3 = 10uL(4h^2+2hH+9h+H)
\end{aligned}
\right..
\end{equation}

Different from naive thinking, $p$ and $t$ do not distribute model memory evenly, part of  activated peak size can only split by tensor parallelism, and the input activation of each layer must be fully loaded based on the tensor parallelism strategy\cite{megatron_lm}.
Therefore, each server must provide a basic memory size to prevent activation memory overflow, which limits the minimum value of $t$.  





\subsection{Search algorithm}

As analyzed above, we impose the following three types of restrictions on the search space:  

\begin{itemize}
\item $G$ can accelerate training, but is specified by the user based on the consideration of the model performance.
\item $b$ affects the computing power utilization and bubble time simultaneously, therefore there exists an optimal value. Based on the white box simulation and memory, we can set an upper limit for the search space of $b$.
\item The choice of $p$ and $t$ is limited by the memory.
\end{itemize}

After pruning, the overall optimal strategy searching algorithm is as Algorithm \ref{alg}.

\begin{algorithm}[h]
\caption{Optimal strategy searching algorithm}
\label{alg}
\textbf{Input}: Hardware conifg, Model config\\
\textbf{Output}: Optimal parallel strategy
\begin{algorithmic}[1] 
\STATE Let $p=1, t=1, b=1$.
\FOR{($p=1 ; p <= N ; ++p$)}
    \FOR{($t=1 ; t<=8 ; t=t*2$)}
        \STATE $b_{max} = f(p,t) $
        \FOR{($b=1 ; b<=b_{max} ; b=b*2$)}
            \IF {OOM}
            \STATE continue.
            \ENDIF
            \STATE $T_T = F(p,t,b,G)$
            \IF{$T_T<T_{min}$}
            \STATE update optimal strategy
            \ENDIF
        \ENDFOR
    \ENDFOR
\ENDFOR
\STATE \textbf{return} optimal strategy
\end{algorithmic}
\end{algorithm}

\section{Experiment and Result}

\subsection{Experimental Setup}

In this section, we conducted experiments to verify the accuracy and robustness of our algorithm.
The main purpose of the experiment was to examine three aspects:

\begin{itemize}
\item Simulation precision: How accurately does the simulation algorithm estimate training time under different parallel settings.
\item Rank preservation: Whether the algorithm can find the global optimal parallel setting and whether the ranking of sub-optimal strategies is accurate.
\item Inference correctness: Whether our inferences about $G, b$, and memory hold.
\end{itemize}

Our experiment was conducted using 16 Ascend 910b NPUs on 2 servers, with each NPU having a maximum computing power of 313T and a memory capacity of 64GB. Internal communication within the server is done through mesh architecture, while communication between servers is done through ring. The total data volume and global batch size of the experiment is fix to 256.
The distributed training framework is ModelLink, a large language models solution that is well adapted to the NPU.
The software environment used is python3.8, pytorch2.1.0. 


\subsection{Simulation precision}

To verify the simulation accuracy of the algorithm for training time, we estimate the training time under different models and parallel configurations, and compare the results with the real profiling. Since most of the profiling tools cannot measure the granularity of our formula \ref{TT_divi}, we combine those item to align with the profiling tools.What needs to be compared is computation duration, communication duration, and the overlap. Among them, bubbles time and reduce computation time are both included in the communication duration.

Table \ref{result_1} shows the simulation precision of our algorithm. 
The experimental models include Baichuan2-7b\cite{yang2023baichuan}, Qwen-14b\cite{bai2023qwen} and Aquila2-7b\cite{zhang2024aquila}.
The number of layers of the model is fixed at 32 to reduce the memory overhead and allow us to compare more combinations.
We not only measured the optimal parallel strategy, but also the top 5 suboptimal strategies to verify the reliability of the algorithm. Strategies are sorted by their throughput.

\begin{table*}[h]
    \centering
    \small
    \setlength{\tabcolsep}{5pt}
    \begin{tabular}{c|c|ccc|ccc|ccc|ccc}
    \hline
        \textbf{Model } &
        \textbf{(d, t, p, b)} & \textbf{\makecell{Total\\Est}} & \textbf{\makecell{Total\\Real}} & \textbf{\makecell{ACC\\(\%)}} & \textbf{\makecell{Cmpt\\Est}} & \textbf{\makecell{Cmpt\\Real}} & \textbf{\makecell{ACC\\(\%)}} & \textbf{\makecell{Comm\\Est}} & \textbf{\makecell{Comm\\Real}} & \textbf{\makecell{ACC\\(\%)}} & \textbf{\makecell{Olap\\Est}} & \textbf{\makecell{Olap\\Real}} & \textbf{\makecell{ACC\\(\%)}} \\ \hline
        ~ & (2 ,4 ,2 ,2) & 25.73  & 27.22  & 94.53  & 20.13  & 19.70  & 97.82  & 8.47  & 10.30  & 82.22  & 2.87  & 2.78  & 96.82   \\ 
        ~ & (2 ,8 ,1 ,2) & 25.88  & 27.96  & 92.55  & 21.49  & 19.70  & 90.92  & 7.69  & 10.26  & 74.96  & 3.30  & 2.79  & 81.80   \\ 
        Baichuan2-7b & (2 ,2 ,4 ,2) & 27.90  & 28.37  & 98.35  & 22.05  & 21.54  & 97.66  & 8.61  & 9.83  & 87.52  & 2.75  & 3.02  & 90.97   \\ 
        ~ & (1 ,8 ,2 ,2) & 28.99  & 29.53  & 98.17  & 23.16  & 22.83  & 98.55  & 9.13  & 10.01  & 91.21  & 3.30  & 3.29  & 99.47   \\ 
        ~ & (1 ,4 ,4 ,2) & 29.43  & 30.11  & 97.76  & 23.04  & 22.53  & 97.74  & 9.27  & 10.52  & 88.06  & 2.87  & 2.98  & 96.51   \\ \hline
        ~ & (2 ,4 ,2 ,2) & 18.09  & 19.02  & 95.13  & 15.05  & 14.55  & 96.54  & 5.25  & 6.71  & 78.16  & 2.21  & 2.20  & 99.52   \\ 
        ~ & (2 ,8 ,1 ,2) & 18.36  & 20.08  & 91.44  & 15.80  & 15.77  & 99.78  & 4.96  & 6.73  & 73.63  & 2.40  & 2.41  & 99.47   \\ 
        Qwen-14b & (2 ,2 ,4 ,2) & 20.33  & 20.59  & 98.75  & 16.88  & 16.20  & 95.80  & 5.58  & 6.50  & 85.88  & 2.13  & 2.10  & 98.77   \\ 
        ~ & (1 ,8 ,2 ,2) & 20.49  & 20.67  & 99.16  & 17.12  & 16.61  & 96.89  & 5.77  & 6.80  & 84.89  & 2.40  & 2.49  & 96.74   \\ 
        ~ & (1 ,4 ,4 ,2) & 20.80  & 21.00  & 99.01  & 17.37  & 16.61  & 95.41  & 5.65  & 6.67  & 84.67  & 2.22  & 2.29  & 97.35   \\ \hline
        ~ & (2 ,8 ,1 ,8) & 3.13  & 3.23  & 96.77  & 2.18  & 2.20  & 98.75  & 1.31  & 1.64  & 79.76  & 0.36  & 0.61  & 58.42   \\ 
        ~ & (2 ,8 ,1 ,4) & 3.20  & 3.30  & 96.88  & 2.26  & 2.31  & 97.78  & 1.31  & 1.60  & 81.50  & 0.37  & 0.62  & 60.29   \\ 
        Aquila2-7b & (4 ,4 ,1 ,4) & 3.25  & 3.35  & 96.80  & 2.16  & 2.18  & 99.36  & 1.44  & 1.75  & 82.36  & 0.36  & 0.57  & 62.35   \\ 
        ~ & (4 ,4 ,1 ,2) & 3.35  & 3.46  & 96.67  & 2.28  & 2.35  & 96.87  & 1.44  & 1.66  & 86.96  & 0.38  & 0.55  & 68.29   \\ 
        ~ & (2 ,8 ,1 ,2) & 3.47  & 3.71  & 93.49  & 2.58  & 2.62  & 98.64  & 1.31  & 1.69  & 77.51  & 0.42  & 0.60  & 70.99   \\ \hline
    \end{tabular}
    \caption{Simulation and precision of training duration for different strategies under 16 NPUs training.}
    \label{result_1}
\end{table*}

It can be seen that our estimation accuracy of the overall computation time is quite good, with an average estimation precision of 96.89\% and a minimum overall accuracy over 91.44\%. This indicates that our modeling of training duration is reliable and provides a solid foundation for determining the optimal strategy. Observing the fitting effect of molecular terms, it can be found that the estimation of computational time is the most accurate. Based on the estimation of computing power utilization, the estimation precision of computation duration can reach an average of 97.45\%.
Estimates of the duration of communication are not very accurate and generally low. This is because the slight deviation in computing efficiency of each worker introduces random waiting times. In our algorithm, the waiting time is about 1 second, and we do not deal with this random term specifically because it has limited influence on optimal strategy planning.

\subsection{Rank preservation}

In the test, our algorithm successfully suggests the global optimal strategy for all the tested models.
The optimal parallel strategy for Baichuan2-7b and Qwen-14b is $(2,4,2,2)$. 
Among the remaining strategies, $t=4$ is optimal, slightly better than $t=8$ suggested by Megatron-LM.

In addition, our algorithm not only accurately finds the global optimal parallel strategy, but also correctly estimates and sorts the top five sub-optimal strategies. 
In the experiment, all the strategies are sorted correctly, which proves that our algorithm has reliable rank-preserving property. 
The deviation between the estimated time and the actual training time is stable.


\subsection{Inference correctness}

In previous section, we propose several pruning theorems based on our white box model. These theorems constrain our searching space, such as $G$ and $b$, enabling enumeration based algorithms to conduct. Previous work often ignores these two hyperparameters, especially in the setting of $b$, where most of the work was controversial. For example, Gavatron sets $b$ to 1, while Calculon sets it to the maximum value within the memory capacity. 
Here, we set up experiments to verify the validity of the inference of $G$ and $b$.

\subsubsection{Training accelerate from $G$.}

\begin{figure}[h]
\centering
\includegraphics[scale=0.55]{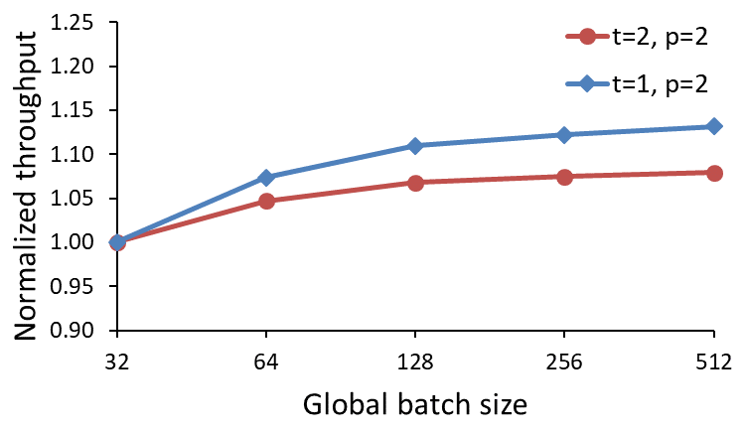}
\caption{The impact of global batch size on throughput.}
\label{gbs_test}
\end{figure}

By observing the single-step training time under different $G$ settings, we evaluate the efficiency of the algorithm in resource utilization. While keeping other parameters unchanged, we gradually doubled $G$ from 32 to 512, and recorded the single step time of training for each $G$.
The result of training efficiency evaluation is shown in Figure \ref{gbs_test}. When $G$ increases from 32 to 512, the number of batches processed per second increases, indicating that the model performs well in utilizing computational resources. A larger $G$ allows for more efficient sample usage, reducing the computational overhead for each iteration, thereby accelerating the training process. However, the acceleration effect of $G$ is only effective for a part of the time-consuming term, and a decrease in the acceleration effect can be observed.

\subsubsection{Optimal value of $b$.}

\begin{figure}[h]
\centering
\includegraphics[scale=0.55]{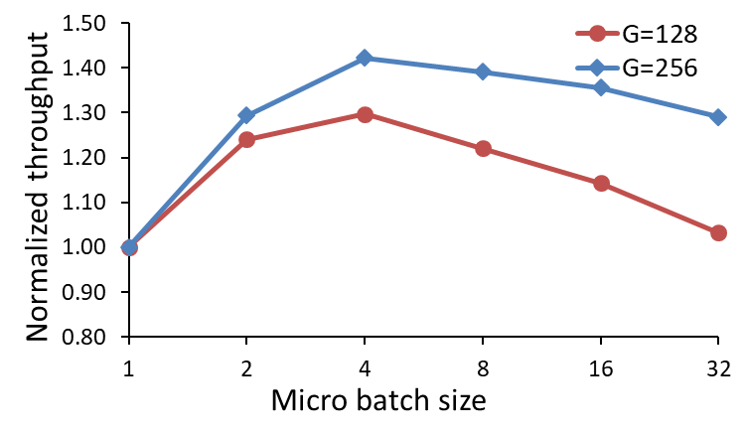}
\caption{The impact of micro batch size on throughput.}
\label{mbs_test}
\end{figure}

For the testing of $b$, we fix $G$ instead. We evaluate the effect of $b$ by observing the single step training time under different $b$ settings. While keeping other parameters constant, we gradually increase $b$ from 1 to 32 at a rate of power 2 and record the single step training time.
The evaluation results are shown in Figure \ref{mbs_test}. The optimal value of $b$ appears at 4, rather than 1 or the maximum capacity of memory. This is consistent with the analysis in Megatron\cite{megatron_lm}. Depending on the model, the optimal value of $b$ is usually between 1 and 4, but it is not easy to determine the specific optimal value. Limiting through white box model alignment and then conducting small-scale enumeration is a feasible strategy.

\section{Conclusion}
In this article, we propose an automatic parallel algorithm based on simulation and strategies search. Our algorithm automatically adjusts the configuration of parallel strategies based on the specifications of the training cluster and models, providing the most efficient training strategies. 
Experimental results have shown that our algorithm can make very accurate estimates of training time. Even though the optimal strategies for different models may vary greatly, our algorithm can accurately find the global optimum.
In recent years, many efforts have been made to address the issue of automatic parallelism. 
The main difficulty lies in the fact that simulating and modeling training duration is a very detailed and tedious task, and some small errors can easily occur and accumulate. 
Moreover, when theoretical simulation models are applied in practice, they inevitably encounter many interferences, which is difficult to quantify , resulting in system instability. 
It is very difficult to establish a precise and robust simulation model. 
In this work, we have made fine adjustments to the simulation model, and many items in the model have undergone multiple iterations to ensure the accuracy of the estimation. 
However, in terms of communication duration simulation, there is still space for improvement in our current work, and we expect to further optimize this part in future iterations.
\clearpage
\bibliography{aaai25}

\end{document}